\documentclass{article} % For LaTeX2e

\usepackage{iclr2024_conference,times}
\usepackage{CJKutf8}
\usepackage{amsmath,amsfonts,bm}

\def\eqref#1{equation~\ref{#1}}
\def\1{\bm{1}}

\DeclareMathAlphabet{\mathsfit}{\encodingdefault}{\sfdefault}{m}{sl}
\SetMathAlphabet{\mathsfit}{bold}{\encodingdefault}{\sfdefault}{bx}{n}

\usepackage{hyperref}
\usepackage{url}
\usepackage{inconsolata}
\usepackage{CJKutf8}
\usepackage{graphicx}
\usepackage{amssymb}
\usepackage{multirow}
\usepackage{pifont}
\usepackage{tabu}
\usepackage{booktabs}
\usepackage{color}
\usepackage{subcaption}
\usepackage{tablefootnote}
\usepackage{wrapfig}
\usepackage{caption}
\usepackage{threeparttable}
\usepackage{colortbl}
\title{KwaiYiiMath: Technical Report}

\author{Jiayi Fu, Lei Lin, Xiaoyang Gao, Pengli Liu, Zhengzong Chen, Zhirui Yang, Shengnan Zhang, \\
\textbf{Xue Zheng, Yan Li, Yuliang Liu, Xucheng Ye, Yiqiao Liao, Chao Liao, Bin Chen, Chengru Song, } \\
\textbf{Junchen Wan$^{\dagger}$, Zijia Lin, Fuzheng Zhang, Zhongyuan Wang, Di Zhang, Kun Gai} \\
\\ \textbf{Kuaishou Technology} \\
}

\iclrfinalcopy % Uncomment for camera-ready version, but NOT for submission.
\begin{document}

\renewcommand{\thefootnote}{$^{\dagger}$}
\footnotetext{Corresponding author: wanjunchen@kuaishou.com.}
\renewcommand{\thefootnote}{\arabic{footnote}}

\maketitle

\begin{abstract}
% Recent advancements in large language models (LLMs) have demonstrated remarkable performance in handling a variety of downstream tasks in the NLP domain.
Recent advancements in large language models (LLMs) have demonstrated remarkable abilities in handling a variety of natural language processing (NLP) downstream tasks, even on mathematical tasks requiring multi-step reasoning. In this report, we introduce the \textbf{KwaiYiiMath} which enhances the mathematical reasoning abilities of KwaiYiiBase\footnote{KwaiYiiBase is a large language model developed by Kuaishou \url{https://github.com/kwai/KwaiYii/}.}, by applying Supervised Fine-Tuning (SFT) and Reinforced Learning from Human Feedback (RLHF), including on both English and Chinese mathematical tasks. Meanwhile, we also constructed a small-scale Chinese primary school mathematics test set (named KMath), consisting of 188 examples to evaluate the correctness of the problem-solving process generated by the models. Empirical studies demonstrate that KwaiYiiMath can achieve state-of-the-art (\textbf{SOTA}) performance on GSM8k, CMath, and KMath compared with the similar size models, respectively.
% There are also many works that show the strong ability of large models to solve mathematical reasoning problems.
% However, most of the existing mathematical LLMs or general LLMs are mainly aimed at solving mathematical calculation problems or English mathematical tasks. 

% To prompt the field of Chinese mathematical tasks, we construct a small-scale Chinese primary school mathematics test set (named KMath), consisting of 188 examples.
% Meanwhile, we also present the \textbf{KwaiYiiMath} which enhances the mathematical reasoning abilities of KwaiYiiBase, by applying Supervised Fine-Tuning (SFT) and Reinforced Language Model Human Feedback (RLHF),
% which is a widely recognized mathematical reasoning benchmark.
% Furthermore, we show the experiment results in a small-scale Chinese primary school mathematics test set to demonstrate that KwaiYiiMath can also achieve SOTA performance on a Chinese mathematical dataset.
\end{abstract}

\section{Introduction}

% Recently, large language models (LLMs) have demonstrated remarkable abilities in handling a variety of natural language processing (NLP) downstream tasks. 
Recent advances in large language models (LLMs) 
% based on the training objective of the next token prediction 
have revolutionized the natural language processing (NLP) landscape~\cite{kenton2019bert,brown2020language}, where scaling up model size and the amount of data is one of the key ingredients~\cite{rae2021scaling,chowdhery2022palm,anil2023palm,touvron2023llama,touvron2023llama2}.
State-of-the-art models trained on vast amounts of data with extremely large model sizes, such as ChatGPT~\cite{openai2022chatgpt}, GPT-4~\cite{openai2023gpt4} and PaLM2~\cite{anil2023palm}, have shown unprecedented performance on a wide range of NLP tasks~\cite{brown2020language,rae2021scaling,du2022glam,lewkowycz2022solving,chowdhery2022palm,ouyang2022training,tay2022ul2,openai2022chatgpt,openai2023gpt4,anil2023palm,touvron2023llama2}.

Surprisingly, recent progress suggests that LLMs also have the potential to solve reasoning problems~\cite{clark2020transformers,talmor2020leap,suzgun2022challenging,wei2022chain}. Specifically, LLMs can perform soft deductive reasoning over natural language descriptions with \textit{implicit knowledge} stored in their parameters~\cite{wei2022chain,kojima2022large,fu2022complexity,shi2022language,zhang2022automatic,zhou2022teaching,diao2023active,shum2023automatic} or \textit{explicit knowledge} in external resources~\cite{wang2022iteratively,creswell2022selection,zhou2022least,press2022measuring,dua2022successive,reppert2023iterated}, and perform step-by-step reasoning just with a few demonstrations or instructions via chain-of-thought prompting (CoT)~\cite{wei2022chain}.
% Surprisingly, recent progress suggests that LLMs also have the potential to solve reasoning problems~\cite{clark2020transformers,talmor2020leap,suzgun2022challenging,wei2022chain}. Specifically, LLMs can perform soft deductive reasoning over natural language descriptions with \textit{implicit knowledge} stored in their parameters~\cite{wei2022chain,kojima2022large,fu2022complexity,shi2022language,zhang2022automatic,zhou2022teaching,diao2023active,shum2023automatic} or \textit{explicit knowledge} in external resources~\cite{wang2022iteratively,creswell2022selection,zhou2022least,press2022measuring,dua2022successive,reppert2023iterated}, and perform step-by-step reasoning just with a few demonstrations or instructions via chain-of-thought prompting (CoT)~\cite{wei2022chain}.

In this report, we focus on how to enhance the mathematical reasoning capabilities of LLM through an alignment process that includes supervised fine-tuning (SFT) and reinforcement learning from human feedback (RLHF).
Specifically, we introduce the KwaiYiiMath which is finetuned with human alignment techniques from KwaiYiiBase to tackle
mathematical problems.

% 实验结果
Experimental results show that KwaiYiiMath outperforms many open-source models in similar sizes by a large margin and is approaching GPT-4 on three mathematical benchmarks including both English and Chinese, i.e., GSM8k~\cite{cobbe2021training}, CMath~\cite{wei2023cmath}, and a small-scale in-house dataset KMath.

% 技术报告的结构
The structure of this report is as follows: Section 2 provides an overview of related work including LLM and LLMs' reasoning ability.
Section 3 introduces the methodology of KwaiYiiMath including the process of supervised fine-tuning and human preference alignment.
Additionally, it also describes details about the efforts in collecting large amounts of mathematical high-quality training data.
In Section 4, we report the experimental results on two public benchmarks and an in-house dataset.
Section 5 concludes this report and points out the future work of KwaiYiiMath.

\section{Related Work}

\paragraph{Large Language Models}
Nowadays, the advent of LLMs encourages the rethinking of the possibilities of artificial general intelligence (AGI). A recent report has even argued that GPT-4 might be \textit{an early version of AGI system}~\cite{bubeck2023sparks}. The success of LLMs consists of three major aspects, including \textit{pre-training} (how to pre-train a LLM based on large amounts of unlabelled data), \textit{adaption} (how to adapt the LLMs for better interaction with humans) and \textit{utilization} (how to use LLMs for solving various downstream tasks)~\cite{zhao2023survey}. By pre-training on large-scale corpora with language modeling objective, LLMs can acquire essential language understanding and generation abilities~\cite{brown2020language,chowdhery2022palm}.\footnote{We omit most details in \textit{pre-training} since KwaiYii Math only concentrates on how to enhance the mathematical reasoning
abilities of KwaiYiiBase.}

However, obviously, one of the major issues is the gap between the training objective and users' objective: users want the model to ``understand and follow their instructions'' while the LLMs are designed to predict the next token. To bridge the gap, instruction tuning (IT) and alignment tuning (AT) are proposed to enhance the capabilities and controllability of LLMs, where IT mainly aims to unlock the abilities of LLMs while the purpose of AT is to align the behaviors of LLMs with human preferences. It refers to the process of further fine-tuning pre-trained LLMs on a collection of formatted instance pairs (i.e., (\textit{instruction}, \textit{output})), where \textit{instruction} and \textit{output} denote the human instruction and the desired output generated by the LLM that follows the human instruction, respectively. Specifically, we first need to collect or construct (\textit{instruction}, \textit{output}) pairs, including manually constructed formatted instances~\cite{mishra2021cross,victor2022multitask,muennighoff2022crosslingual,wang2022super,longpre2023flan,zhou2023lima,conover2023free,kopf2023openassistant}, and automatically constructed formatted instances~\cite{wei2021finetuned,bach2022promptsource,honovich2022unnatural,wang2022self,xu2023wizardlm,xu2023baize,ji2023towards}. Note that it has been widely shown that the number of tasks, the quality and diversity of instruction instances poses an important impact on the performance of LLMs~\cite{ouyang2022training,victor2022multitask,wei2021finetuned,wang2022super,chung2022scaling,taori2023stanford,zhou2023lima}. Then, we use these formatted instances or carefully selected formatted instances to further fine-tune LLMs in a supervised learning way (also known as SFT) according to different requirements. For example, we can use formatted instances of different modality, domains and applications to obtain different specific LLMs, including multimodal LLMs (InstructPix2Pix~\cite{brooks2023instructpix2pix}, LLaVA~\cite{liu2023visual}, Video-LLaMA~\cite{zhang2023video}, InstructBLIP~\cite{dai2023instructblip}, MultiModal-GPT~\cite{gong2023multimodal}), and domain and application specific LLMs (InstructDial~\cite{gupta2022instructdial}, LINGUIST~\cite{rosenbaum2022linguist}, InstructUIE~\cite{wang2023instructuie}, Writing-Alpaca~\cite{zhang2023multi}, Radiology-GPT~\cite{liu2023radiology}, ChatDoctor~\cite{yunxiang2023chatdoctor}, Goat~\cite{liu2023goat}, WizardCoder~\cite{luo2023wizardcoder}, etc). However, the SFT process seems not to be stable enough due to the small amount of data and large model size. So, some studies focus on how to combine instruction tuning and pre-training, mainly consisting of two directions: a one-stage process (pre-trained from scratch with a mixture of pre-training data and instruction tuning data)~\cite{raffel2020exploring,zeng2022glm} and two-stage process (fine-tuned with a mixture of pre-training data and instruction tuning data)~\cite{iyer2022opt}. AT refers to the process of reinforcement learning from human feedback (RLHF) for better aligning LLMs with human preferences~\cite{ouyang2022training}. RLHF mainly comprises three key components: an LLM after SFT, a reward model learning to reflect human feedback for the text generated by the LLM, and a reinforcement learning (RL) algorithm (e.g., Proximal Policy Optimization (PPO)~\cite{schulman2017proximal}) to align the LLM based on the guidance signals generated by the reward model. 
We train a reward model to predict the human-preferred output based on the collected human feedback data and align the LLM based on the reward signals generated by the reward model using RL algorithm.\footnote{We omit the details of alignment criteria and human feedback collection for brevity.}

After pre-training and adaption, LLMs can serve as a general-purpose language task solver (to some extent) by simply conditioning the models on a few examples (few-shot) or instructions describing the task (zero-shot). The success of LLMs is often attributed to few-shot (in-context) or zero-shot learning (i.e., emergent abilities~\cite{wei2022emergent} that may not be observed in previous smaller language models).\footnote{Note that emergent abilities may not occur in some LLMs.} This leads to the rapid development of the ``prompting'' technique, revolutionizing the way that humans develop and use AI algorithms. Thus, designing prompts has become a hot topic in NLP, including demonstration selection~\cite{liu2021makes,rubin2021learning,xie2021explanation,zhang2022active,kim2022self,lee2022does,levy2022diverse,su2022selective,ye2022complementary}, and demonstration order~\cite{liu2021makes,lu2021fantastically}.

\paragraph{Reasoning with Large Language Models}
Reasoning, the process of making inferences based on existing knowledge, is the core of human intelligence. However, existing LLMs have struggled to achieve high performance on \textit{system-2} tasks requiring slow and multi-step reasoning such as mathematical and commonsense reasoning~\cite{rae2021scaling}. To enhance the reasoning ability of LLMs prompting, there are two major directions: strategy-enhanced reasoning and knowledge-enhanced reasoning~\cite{qiao2022reasoning}. Strategy-enhanced reasoning refers to design a better reasoning strategy, such as the prompt design in single reasoning stage~\cite{wei2022chain,kojima2022large,fu2022complexity,shi2022language,zhang2022automatic,zhou2022teaching,diao2023active,shum2023automatic}, reasoning stage by stage~\cite{wang2022iteratively,creswell2022selection,zhou2022least,press2022measuring,dua2022successive,reppert2023iterated}, natural language rationales optimization~\cite{ye2022unreliability,wang2022self,huang2022large,li2023making,weng2023large,yoran2023answering,shinn2023reflexion,madaan2023self,paul2023refiner}, and external engine augmentation~\cite{liu2022mind,chen2022program,gao2023pal,lyu2023faithful,imani2023mathprompter}. Knowledge enhanced reasoning refers to prompt LLMs with \textit{implicit} knowledge stored in LLM~\cite{li2022explanations,wang2022pinto,magister2022teaching,ho2022large,fu2023specializing} or \textit{explicit} knowledge in external resources~\cite{yang2022logicsolver,su2022selective,lu2022dynamic,he2022rethinking}, such as wiki, the reasoning steps, etc.

\section{Method}
\begin{figure*}[!t]
    \centering
    \includegraphics[width=1.0\linewidth]{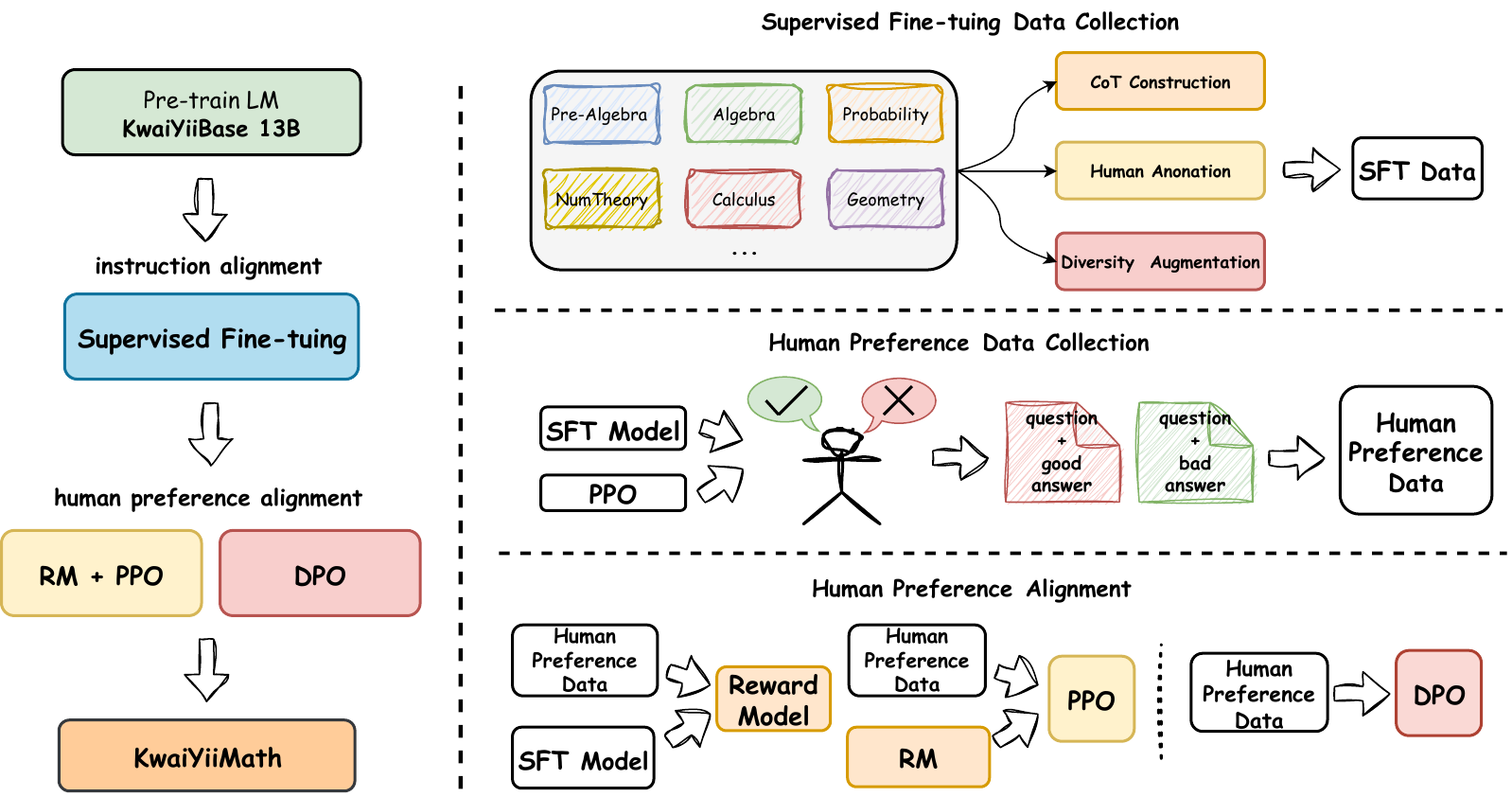}
    \caption{The overview of KwaiYiiMath. The left part is the framework of the model training process. The right part shows the details of three main components, including the SFT data collection, human preference data collection, and human preference alignment training. The data used by RM, PPO, and DPO are only in the same form and collection method, but their training datasets are different.}
    \label{fig:framework}
\end{figure*}

In this section, we introduce the details of KwaiYiiMath.
Figure~\ref{fig:framework} shows the overview of our model.
Specifically, the left part of Figure~\ref{fig:framework} shows the training process overview of KwaiYiiMath, and it mainly consists of two steps, supervised fine-tuning and human preference alignment.
 % Following the training paradigm of InstructGPT~\cite{ouyang2022training}, we first use instruction-response pair data to fine-tune KwaiYiiBase in a supervised fashion.
% do supervised fine-tuning on the pre-trained model.
The right part of Figure~\ref{fig:framework} shows the main three components, including SFT data collection, human preference data collection, and human preference alignment training.
We first collect high-quality mathematical instruction data, in the form of $<$\textit{question, answer}$>$ pairs, to do the supervised fine-tuning.
Next, for each question, we first generate \emph{K} different answers from SFT models and actor models from reinforcement learning respectively, then classify them into good answers and bad answers with the help of human annotation.
In this way, we can collect a large amount of high-quality data representing human preference and then use this data to train the reward model and do human preference alignment training.

% they are not the same training set.

\subsection{Supervised fine-tuning}
Previous work has shown that the diversity and quality of instruction data poses an important impact on the SFT performance~\cite{zhou2023lima,peng2023instruction,chen2023alpagasus}, a conclusion that also holds true in LLMs' mathematical reasoning ability~\cite{yuan2023scaling,luo2023wizardmath,abel,yu2023metamath}. Therefore, we focus on how to collect or construct high-diversity and high-quality instruction data. 

To obtain as much diversity as possible in math data, we first collect math data from a wide range of sources, including different difficulties (e.g., primary school, middle school, and university, etc.), and different fields of math (e.g., algebra, geometry, and probability, etc.). 
Then, we generate intermediate rationales for math questions only with the final answer or without the answer using open-source LLMs and ensure the correctness of intermediate rationales and answers through manual annotation.
We try to construct intermediate rationales for all mathematical instruction data since Chain-of-Thought (CoT)~\cite{wei2022chain} has been proven effective either in prompting or instruction data for fine-tuning~\cite{ho2022large,zhu2022solving,li2023symbolic}. 
In addition to mathematical data, we also sample 300k open-domain conversations from KwaiYiiChat\footnote{KwaiYiiChat is the finetuned model from KwaiYiiBase for open-domain conversations.} training data to maintain the model's ability to handle open-domain questions.
More details of SFT data collection are as follows.

\subsubsection{Data Diversity}
\begin{figure*}[!t]
    \centering
    \includegraphics[width=1.0\linewidth]{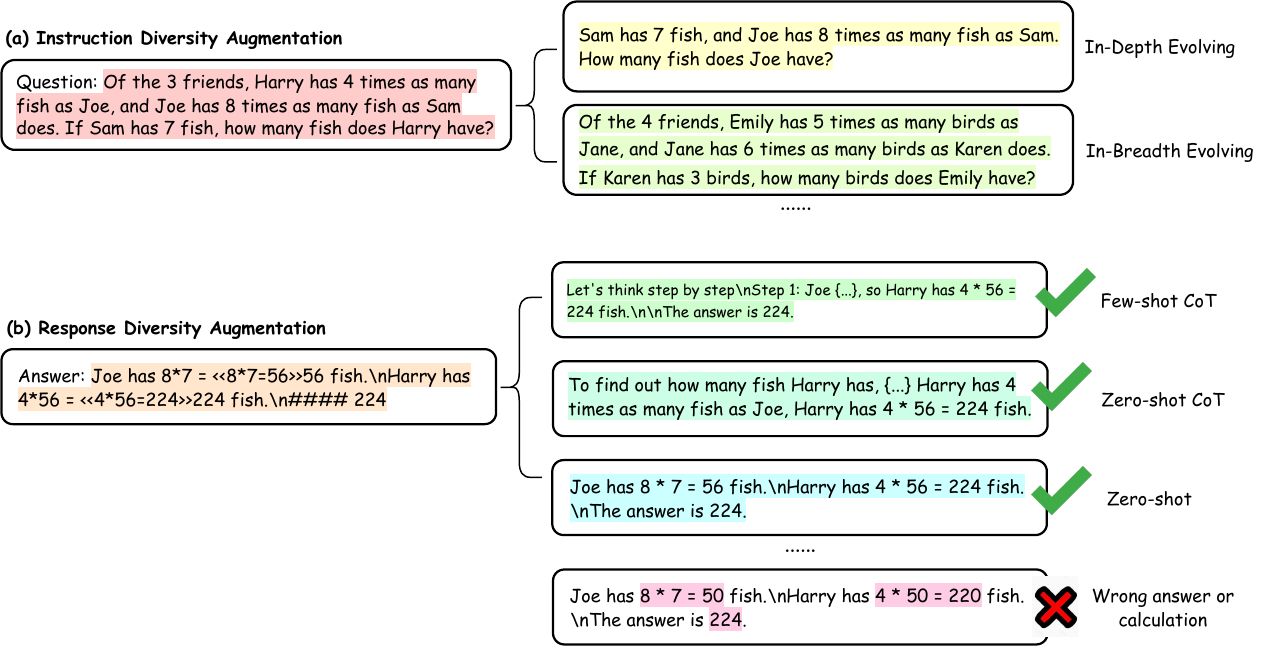}
    % \caption{The construction process of diverse instructions and responses.}
    \caption{An example of SFT data diversity augmentation. We chose a sample from the train set of GSM8k. The upper part shows the instruction diversity augmentation process. There are two new instructions that are generated through the in-depth evolving and in-breadth evolving, respectively. The bottom part shows the response diversity augmentation process. There are four new responses that are generated through diverse reasoning path construction and one of them is filtered out due to wrong calculation, even the final answer is correct.}
    \label{fig:fig1}
\end{figure*}

% 数据多样性包括两方面：问题的多样性和回答的多样性。
We consider the diversity of instruction data mainly from two aspects: the diversity of instructions and responses, respectively. Figure~\ref{fig:fig1} illustrates the construction process of SFT data with an example. 
% the diversity of responses.
% The diversity of instruction data mainly includes two aspects: diversity of instructions and diversity of responses.

% \subsubsection*{Instruction Diversity}
% 在提升指令多样性方面
\paragraph{Instruction Diversity} Inspired by Evol-Instruct~\cite{xu2023wizardlm, luo2023wizardcoder} which uses LLM instead of humans to generate diverse instructions through a manual designed set of evolutionary actions, we designed Dual-evolving actions in depth and Constrained-mutation evolving actions in breadth for math instruction.
Nearing completion of our work, the authors of Evol-Instruct~\cite{xu2023wizardlm, luo2023wizardcoder} had also adapted their idea to the math LLM and further proposed Reinforcement Learning from Evol-Instruct Feedback~\cite{luo2023wizardmath} which combines Evol-Instruct with reinforcement learning.
However, the Evol-Instruct actions we use are somewhat different. 
% 和wizardmath的区别就是，他们的指令降级是一对一的 我们是一对多 优点：数据量翻倍 数据域范围更大
% We define the data is $\mathcal{D}$, $P_{\mathcal{D}}$ denotes the data distribution.
Specifically, we consider a logically complex mathematical problem $Q$ is constructed by a series of simple sub-problems $[q_1,q_2,\ldots,q_n]$. 
The original problem $Q$ naturally has a CoT relation within sub-problems $\{q_i\}_{i=1\ldots n}$. 
The math-solving process can be considered as the CoT process of these sub-problems. 
Therefore, how to retain the CoT process in Evol-Instruct is a matter. 
Evol-Instruct mentioned in~\citet{luo2023wizardmath} evolves original instruction $Q$ to $Q'$, which is a one-to-one paradigm no matter Downward evolution or Upward evolution action and will hurt the CoT process of $Q$ when making Downward evolution especially. 
We propose Dual-evolving actions in depth to retain the CoT process and expand instruction diversity in the meantime. 
Specifically, Dual-evolving actions include two steps:
\begin{itemize}
    \item First we design prompts to decompose a mathematical problem $Q$ into multiple sub-problems $[q_1,q_2,\ldots,q_n]$. Each sub-problem is individual from each other, and the CoT process consists of the order of sub-problems, which is a one-to-many paradigm of Downward evolution.
    \item Second we design prompts to increasingly enhance the difficulty of sub-problems and further improve the diversity of instructions. Since our method consists of two steps, we call it a Dual-evolving action.
\end{itemize}
% 
% Therefore, the ability of LLM to abstract sub-problems from a complex math problem is essential in math reasoning. In order to empower LLM with this ability, we employ dual-evolving actions in math instruction. 
%

% Specifically, in the in-depth evolving, we adopt a method Dual-evolving actions and design prompts to decompose a mathematical problem into multiple sub-problems that can be solved incrementally. The difference with Evol-Instruct actions mentioned in\cite{luo2023wizardmath} is that our method keep the COT process of original math problem and is one-to-many evolving action. 

In the in-breadth evolving, we use a constrained-evolving action that is inspired by the mutation evolution~\cite{xu2023wizardlm}. 
% Specifically, we designed a prompt to evolve a new problem from an existing mathematical problem in a 
Specifically, We designed a prompt to evolve new problems based on existing problems within a constrained scope.
The purpose of adding scope constraints is to avoid evolving actions that lead to unsolvable mathematical problems.
% which involves prompt construction of a similar new problem based on a mathematical problem.

% todo: 需要完善, 补充更加详细的内容, 具体广度上的动作是怎么做的，增加篇幅

% \subsubsection*{Rsesponse Diversity}
\paragraph{Response Diversity}
% 在提升response多样性方面 - reasoning paths WIP
Previous work~\cite{ho2022large,yuan2023scaling} has shown that diverse reasoning paths can improve the reasoning performance of LLM, including: for a given instruction sample in a training set, using multiple different models or sampling strategies to generate multiple reasoning paths.
% (i) using a sampling strategy to generate multiple reasoning paths; (ii) using multiple different models to generate multiple reasoning paths.

% 大致思路
Inspired by that, we collect various available LLMs including open-source LLMs such as Llama~\cite{touvron2023llama,touvron2023llama2} of different sizes, etc and different versions of KwaiYiiMath. Then, we use these models to generate multiple reasoning paths given each question. 
Specifically, we fine-tune open-source LLMs on mathematical datasets in a supervised fashion to 
obtain the ability to generate more correct reasoning paths.\footnote{Different versions of KwaiYiiMath are already fine-tuned on mathematical datasets in a supervised fashion.} To further augment such abilities of LLMs and improve the diversity of reasoning paths, we collect diverse CoT prompts,\footnote{https://github.com/FranxYao/chain-of-thought-hub/tree/main/gsm8k/lib\_prompt} and use different prompting strategies such as zero-shot, zero-shot CoT and few-shot CoT. For each question $q_{i}$, we generate $k$ candidate reasoning paths and answers $r$, $a$ with a temperature of 0.7 following~\citet{cobbe2021training} and one of prompting strategies based on different prompts. We first filter out reasoning paths with wrong answers $a \neq a_{i}$ or wrong calculations where equations are extracted from reasoning paths.\footnote{Note that we only evaluate the correctness of extracted complete equation (e.g., 3 $+$ 4 $=$ 7) instead of incomplete equation (e.g., $+$ 4 $=$ 7).} We retain all reasoning paths with the same equation list as the augmented data unlike~\cite {yuan2023scaling} since we argue that diverse contexts also pose an important impact on the reasoning performance of LLMs.
% Finally, we fine-tune KwaiYiiBase on the augmented data in a supervised fashion.

% To generate as many diverse reasoning paths as possible, we also construct a collection of few-shot prompts to guide the LLM to generate highly diverse reasoning paths for each math question in the training set~\cite{li2023making}. 
% Furthermore, inspired by related work~\cite{li2023making}, we constructed a collection of few-shot prompts to guide the LLM to generate highly diverse reasoning paths for each math question in the training set. 
% Therefore, we first collect a model set, including open source models such as Llama\cite{touvron2023llama}\cite{touvron2023llama2} of different sizes, as well as different versions of KwaiYiiMath, to generate reasoning paths for samples. 
% For each model in the set, we use it to generate multiple reasoning paths with a repetition penalty.

% 更多的prompt多样性
% To generate as many diverse reasoning paths as possible.
% Furthermore, inspired by related work~\cite{li2023making}, we constructed a collection of few-shot prompts to guide the LLM to generate highly diverse reasoning paths for each math question in the training set. 

% \subsubsection{Data Quality Control}
\subsubsection{Data Quality}
% For the quality mathematical data, the most important thing is the correctness of the calculation process and the correctness of the final answers in the response.
The most important thing for mathematical data is the correctness of the calculation process and the final answers of the response.
We denote $\mathcal{D}=\{q_i, p_i,a_i\}_i$ is the mixed datasets, where $q_i, p_i=\{e_1,e_2,\ldots,e_n,\hat{a}_i\}$ are a question and a reasoning path respectively, $\{e_i\}_i$ denote equation set, $\hat{a}_i$ is the final answer, and $a_i$ denote the ground truth reasoning answer of $q_i$. LLMs often make calculation or conclusion mistakes in the reasoning path~\cite{gao2023pal,chen2022program} such as $eval(e_i)=False$ or $\hat{a}_i\neq a_i$. 

In order to attain high-quality data, we make an effort to ensure the correctness of both the final answers and calculation processes. Specifically, we first use LLM-generated responses to extract the final answers $\hat{a}_i$ and then filter out reasoning paths $p_j$ with wrong answers $p_j:\{\hat{a}_j\neq a_j\}$. Then, we use the regular expression to extract the equation set $\{e_i\}_i$ in response and utilize a Python interpreter to evaluate the correctness of the response. As for the single true answer queries, we control the both correctness of the final answer and the calculation process. As for the multiple true answer queries, we only control the correctness of the calculation process. 

% For the correctness of the final answers, we first use LLM responses to extract the final answers $\hat{a}_i$ and then filter out reasoning paths $p_j$ with wrong answers $p_j:\{\hat{a}_j\neq a_j\}$.

% % 对于单个固定答案的数据, 可以判断答案的正确性和过程计算的正确性; 对于非单独固定答案的数据, 则只做过程计算正确性的判断:
% For the correctness of the calculation process, we first use a regular expression to extract the equation set $\{e_i\}_i$ in response and utilize a Python interpreter to evaluate the correctness of the response. As for the single true answer queries, we control the both correctness of the final answer and the calculation process. As for the multiple true answer queries, we only control the correctness of the calculation process.

\subsection{Human Preferences Alignment}
Despite the significant performance of fine-tuned LLMs on mathematical reasoning abilities, they are still prone to generate content that contains reasoning errors, incorrect answers, or redundant inference processes. We argue that the performance enhancement of LLMs not only derives from supervised fine-tuning but also from human preferences alignment methods. Therefore, we use two scalable alignment frameworks, reinforcement learning from human feedback (RLHF) and DPO~\cite{rafailov2023direct}, both are learning from human preferences for training aligned language models and improving mathematical reasoning abilities and answer correctness.

\subsubsection{Reinforcement Learning From Human Feedback}
RLHF is to apply reinforcement learning directly on LLMs with human preferences as feedback, and developing rapidly in the language models alignment field. Inspired by InstructGPT, we propose a classic RLHF training pipeline that consists of two phases: 1) human preferences comparison data collection and reward model training; 2) reinforcement learning with PPO.

\textbf{Reward Model}: The reward model is used to evaluate the quality of the SFT  generation from the aspect of mathematical result and procedure, as well as human preference.

Given an input, we sample a pair of responses from our SFT and PPO of different versions so that RM can capture diverse data distribution. Some existing open-source preference datasets are also combined to improve the generalization of RM, such as Anthropic Helpful and Harmless~\cite{bai2022training}, OpenAI WebGPT~\cite{nakano2021webgpt}. 
 
The binary ranking loss function we use is consistent with~\cite{ouyang2022training}. We held out 5,000 examples as a test set to evaluate our model. The results are reported in Table~\ref{table:rm}. As a reference point, we also evaluate other publicly available alternative solutions as baselines: the Open Assistant reward model based on DeBERTa V3 Large~\cite{he2020deberta}, and SteamSHP-XL~\cite{ethayarajh2022understanding} based on FLAN-T5-xl.

\textbf{PPO}: PPO is the most famous reinforcement learning method used in RLHF. It utilizes scores from reward models as human feedback signals to fine-tune LLMs. By randomly sampling prompts from SFT datasets and using policy-generated responses, we aim to enable fine-tuned models aligned with human values. We also find it is crucial to whiten reward scores because of the reward hacking issue and increase the training stability.

% 如果reward model详细介绍了q+a多样性过滤策略，这里可以不写了
\subsubsection{Direct Preference Optimization (DPO) }
RLHF methods need a reward model to fit with human preferences, then optimize language models with reinforcement learning to generate high reward score answers. However, they require a large consumption of computational resources and complex large-scale distributed settings. The number of hypermeters in RLHF methods also increases the difficulty of the stability of optimizing results. 

Direct Preference Optimization (DPO) is an RL-free method and easy to implement. By eliminating the reward model, DPO only considers the policy model and reference model probability distributions and uses a simple binary cross-entropy objective to optimize language models from human preferences. DPO training data has a consistent pattern with the reward model data used for training. To effectively improve the mathematical reasoning abilities of KwaiYiiMath, we sample a subset of reward model data and conduct DPO training on the fine-tuned model.

\section{Experiments}
% In this section, we introduce the overview and related details of the experiments.
We mainly evaluate KwaiYiiMath on three comprehensive and realistic benchmarks for measuring mathematical reasoning ability, including two public benchmarks (GSM8k~\cite{cobbe2021training} and CMath~\cite{wei2023cmath}), and an in-house dataset (KMath).
% English and Chinese mathematical datasets ( KMath).

\begin{wraptable}{r}{0.6\linewidth}
\centering
\vspace{-1.0em}
\caption{Results on our Chinese and English test set of human preference benchmarks. }
\label{table:rm}
\resizebox{1.0\linewidth}{!}{
\begin{tabular}{@{\hspace{5pt}}ccc}
\toprule
  RM Model & \bf KwaiYiiMath-en & \bf KwaiYiiMath-zh \\
\midrule
  Open Assistant & 63.79 & 65.80 \\
  SteamSHP-XL    & 55.90 & 54.43 \\
\midrule
  KwaiYiiMath-RM & \bf 77.30 (+13.51) & \bf 78.48 (+12.68) \\
\bottomrule
\end{tabular}
}
\vspace{-0.7em}
\end{wraptable}
\subsection{Evaluation Datasets}
% We mainly evaluate KwaiYiiMath on two benchmarks: GSM8k\cite{cobbe2021training} and KMath. 
GSM8k~\cite{cobbe2021training} contains 7,473 training examples and 1,319 test examples, mainly on grade school-level English math problems. Each question consists of basic arithmetic operations (addition, subtraction, multiplication, and division), and generally requires 2 to 8 reasoning steps to solve. 

CMath~\cite{wei2023cmath} is a Chinese elementary school math word problems dataset that comprises 1.7k\footnote{The CMath data set contains a total of 1.7k data, of which 960 are currently available for download.} elementary school-level math word problems with detailed annotations, sourced from actual Chinese workbooks and exams. 
CMath also has fine-grained annotations, including grade, number of reasoning steps, digits, and distractors.
These annotations can be used to evaluate the LLM's fine-grained mathematical reasoning ability and robustuess mathematical reasoning ability.

KMath is a small-scale in-house mathematical dataset that contains 188 Chinese math questions.
The questions of KMath are mainly on the grade school level, which consists of algebra, calculus, geometry, and probability.

% \begin{wraptable}{r}{1\linewidth}
% \begin{table}[!t]
% \centering
% \small
% \resizebox{0.9\linewidth}{!}{
% \begin{tabular}{c|c c c c}
% \hline
%   Model & \#params & GSM8k & CMath & KMath \\
% \hline
%   GPT4~\cite{openai2023gpt4} & - & $92.00^*$ & $86.00^*$ & 75.00 \\
%   ChatGPT~\cite{openai2022chatgpt} & - & $74.90^*$ & $73.83^*$ &  59.57 \\
%   Ernie Bot & - & 56.23 & 86.83 & 72.87 \\
%   BaiChuan1 & 13B & $26.76^*$ & 51.33 & 28.19 \\
%   WizardMath~\cite{luo2023wizardmath} & 13B & $63.90^*$ & 50.83 & 23.40 \\
%   QWen & 7B & $51.60^*$ & 63.16 & 44.15 \\
%   ChatGLM2 & 6B & $29.20^*$ & 68.36 & 50.00\\
% \hline
% KwaiYiiMath-SFT & 13B & 64.60 & 84.66 & 86.17 \\
% KwaiYiiMath-RLHF & 13B & \textbf{67.80 (+3.9)} & - & \textbf{88.30 (+13.3)} \\
% \hline
% \end{tabular}
% }
% \caption{Results of pass@1 (\%) on GSM8k, CMath and KMath. The character $*$ denotes that results are attained from the related works, and the remaining results are attained from our tests.}
% \label{table:performance}
% \end{table}
\begin{table*}[t]
  \centering
  \caption{Results of pass@1 (\%) on GSM8k, CMath and KMath. The character $*$ denotes that results are attained from the related works, and the remaining results are attained from our tests. The character ${\dagger}$ shows our best result from different human preferences alignment experiments.
  }
  \label{table:performance}
  \begin{tabular}{@{}lccccc@{}}
      \toprule
      \bf Model & & \bf \#Params & \bf GSM8K & \bf CMath & \bf KMath \\
      \midrule
      \multicolumn{6}{c}{\textit{Closed-source models}} \\
      \midrule
     
      \multicolumn{2}{@{}l}{GPT-4~\cite{openai2023gpt4}}  & - & $92.00^*$ & 86.00 & 75.00 \\

      \arrayrulecolor{lightgray}\midrule
      \multicolumn{2}{@{}l}{ChatGPT~\cite{openai2022chatgpt}}  & - & $74.90^*$ & 73.83 & 59.57\\
      
      % \arrayrulecolor{lightgray}\midrule
      % \multicolumn{2}{@{}l}{}  & 8B & $16.20^*$ & - & - \\
      % \multicolumn{2}{@{}l}{Minerva~\cite{lewkowycz2022solving}}  & 62B & $52.40^*$ & - & - \\
      % \multicolumn{2}{@{}l}{}  & 540B & $58.80^*$ & - & - \\

      \arrayrulecolor{lightgray}\midrule
      \multicolumn{2}{@{}l}{\multirow{3}{*}{Minerva~\cite{lewkowycz2022solving}}} & 8B & $16.20^*$ & - & - \\ \multicolumn{2}{@{}l}{} & \color[HTML]{9B9B9B}62B & \color[HTML]{9B9B9B}$52.40^*$ & \color[HTML]{9B9B9B}- & \color[HTML]{9B9B9B}- \\  \multicolumn{2}{@{}l}{} & \color[HTML]{9B9B9B}540B & \color[HTML]{9B9B9B}$58.80^*$ & \color[HTML]{9B9B9B}- & \color[HTML]{9B9B9B}- \\
      
      \arrayrulecolor{lightgray}\midrule
      \multicolumn{2}{@{}l}{Ernie Bot~\cite{ernie2023}} & - & 56.23 & 84.33 & 72.87\\
      
      \arrayrulecolor{lightgray}\midrule
      % \multicolumn{2}{@{}l}{MathQwen~\cite{qwen2023qwen}} & 14B & $69.80^*$ & - & - \\
    \multicolumn{2}{@{}l}{\multirow{2}{*}{MATH-QWEN-CHAT~\cite{qwen2023qwen}}} & 7B & $62.50^*$ & - & - \\ \multicolumn{2}{@{}l}{} & \color[HTML]{9B9B9B}14B & \color[HTML]{9B9B9B}$69.80^*$ & \color[HTML]{9B9B9B}- & \color[HTML]{9B9B9B}- \\
      
      \arrayrulecolor{black}\midrule
      \multicolumn{6}{c}{\textit{Open-source models}} \\
      \arrayrulecolor{black}\midrule

      \multicolumn{2}{@{}l}{\multirow{2}{*}{LLaMA-1~\cite{touvron2023llama}}} & 13B & $17.80^*$ & - & - \\ \multicolumn{2}{@{}l}{} & \color[HTML]{9B9B9B}33B & \color[HTML]{9B9B9B}$35.60^*$ & \color[HTML]{9B9B9B}- & \color[HTML]{9B9B9B}- \\ 
      \arrayrulecolor{lightgray}\midrule
      
      \multicolumn{2}{@{}l}{\multirow{2}{*}{LLaMA-2~\cite{touvron2023llama2}}} & 13B & $28.70^*$ & - & - \\ \multicolumn{2}{@{}l}{} & \color[HTML]{9B9B9B}34B & \color[HTML]{9B9B9B}$44.20^*$ & \color[HTML]{9B9B9B}- & \color[HTML]{9B9B9B}- \\ 
      \arrayrulecolor{lightgray}\midrule
      
      % \multicolumn{2}{@{}l}{BaiChuan1~\cite{baichuan2023baichuan2}} & 13B & $26.76^*$ & 51.33 & 28.19\\

      \multicolumn{2}{@{}l}{\multirow{2}{*}{BaiChuan1~\cite{baichuan2023baichuan2}}} & 7B & $9.17^*$ & - & - \\ \multicolumn{2}{@{}l}{} & 13B & $26.76^*$ & 51.33 & 28.19 \\ 
      \arrayrulecolor{lightgray}\midrule
      
      % \multicolumn{2}{@{}l}{BaiChuan2~\cite{baichuan2023baichuan2}} & 13B & $52.77^*$ & - & -\\
        \multicolumn{2}{@{}l}{\multirow{2}{*}{BaiChuan2~\cite{baichuan2023baichuan2}}} & 7B & $24.49^*$ & - & - \\ \multicolumn{2}{@{}l}{} & 13B & $52.77^*$ & - & - \\ 
      
      \arrayrulecolor{lightgray}\midrule
      % \multicolumn{2}{@{}l}{WizardMath~\cite{luo2023wizardmath}} & 13B & $63.90^*$ & 50.83 & 23.40\\

      \multicolumn{2}{@{}l}{\multirow{2}{*}{WizardMath~\cite{luo2023wizardmath}}}  & 13B & $63.90^*$ & 50.83 & 23.40 \\ \multicolumn{2}{@{}l}{} & \color[HTML]{9B9B9B}70B & \color[HTML]{9B9B9B}$81.60^*$ & \color[HTML]{9B9B9B}- & \color[HTML]{9B9B9B}- \\
      
      \arrayrulecolor{lightgray}\midrule
      \multicolumn{2}{@{}l}{ChatGLM2~\cite{zeng2022glm}} & 6B & $29.20^*$ & 68.36 & 50.00\\
      \arrayrulecolor{lightgray}\midrule
      \multicolumn{2}{@{}l}{QWen~\cite{qwen2023qwen}} & 7B & $51.60^*$ & 63.16 & 44.15\\
      \arrayrulecolor{lightgray}\midrule
      % \multicolumn{2}{@{}l}{GAIRMath-Abel~\cite{abel}} & 13B & $66.40^*$ & - & -\\ 

    \multicolumn{2}{@{}l}{\multirow{3}{*}{GAIRMath-Abel~\cite{abel}}}  & 7B & $59.74^*$ & - & - \\ \multicolumn{2}{@{}l}{} & 13B & $66.41^*$ & - & - \\ \multicolumn{2}{@{}l}{} & \color[HTML]{9B9B9B}70B & \color[HTML]{9B9B9B}$83.62^*$ & \color[HTML]{9B9B9B}- & \color[HTML]{9B9B9B}- \\ 
      
      \arrayrulecolor{lightgray}\midrule
      % \multicolumn{2}{@{}l}{MetaMath~\cite{yu2023metamath}} & 13B & $72.30^*$ & - & -\\ 

     \multicolumn{2}{@{}l}{\multirow{3}{*}{MetaMath~\cite{yu2023metamath}}}  & 7B & $66.50^*$ & - & - \\ \multicolumn{2}{@{}l}{} & 13B & $72.30^*$ & - & - \\ \multicolumn{2}{@{}l}{} & \color[HTML]{9B9B9B}70B & \color[HTML]{9B9B9B}$82.30^*$ & \color[HTML]{9B9B9B}- & \color[HTML]{9B9B9B}- \\

     \arrayrulecolor{black}\midrule
      \multicolumn{2}{@{}l}{\bf KwaiYiiMath} & 13B & 72.33 & 85.33 & 73.40 \\ 
      \multicolumn{2}{@{}l}{\bf KwaiYiiMath-HPA$^{\dagger}$} & 13B & \textbf{73.31} & \textbf{85.83} & \textbf{74.47} \\ 
      \arrayrulecolor{black}\bottomrule
  \end{tabular}
\end{table*}
% \end{wraptable}
\subsection{Baselines}
The baseline models compared in our experiments can be divided into two categories: closed-source models and open-source models.
\paragraph{Closed-source models} Many technology companies have trained LLMs with strong abilities in many downstream tasks, but for some reason do not release their model weights and these models are referred to as closed-source models.
In our experiments, we consider five closed-source LLMs including GPT-4~\cite{openai2023gpt4}, ChatGPT~\cite{openai2022chatgpt}, Ernie Bot\footnote{\url{https://yiyan.baidu.com/}}, Minerva~\cite{zhu2022solving} and MATH-QWEN-CHAT~\cite{qwen2023qwen}. 
\paragraph{Open-source models} There are also many teams that open-source the LLMs they trained, and we can directly download the model weights through open-source communities.
These open-source models also demonstrate excellent capabilities in many downstream tasks.
In our experiments, we select LLaMa1\&2~\cite{touvron2023llama, touvron2023llama2}, BaiChuan1\&2~\cite{baichuan2023baichuan2}, ChatGLM2-6B\footnote{\url{https://github.com/THUDM/ChatGLM2-6B}}, QWen\footnote{\url{https://github.com/QwenLM/Qwen-7B}}, WizardMath\cite{luo2023wizardmath}, GAIRMath-Abel~\cite{abel} ,and MetaMath~\cite{yu2023metamath} as the open-source baselines.

\begin{figure}[!t]
    \centering
    \begin{subfigure}[b]{0.45\textwidth}
     \includegraphics[width=\textwidth]{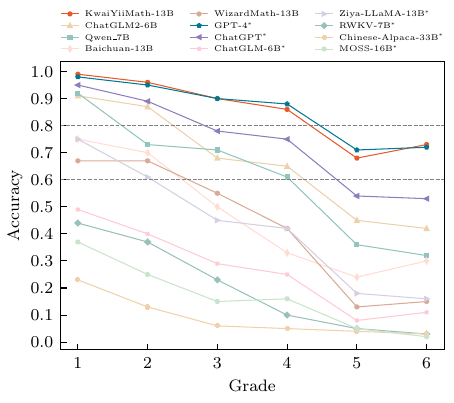}
     \caption{}
     \label{subgraph:a}
    \end{subfigure}
\hspace{-2mm}
    \begin{subfigure}[b]{0.45\textwidth}
     \includegraphics[width=\textwidth]{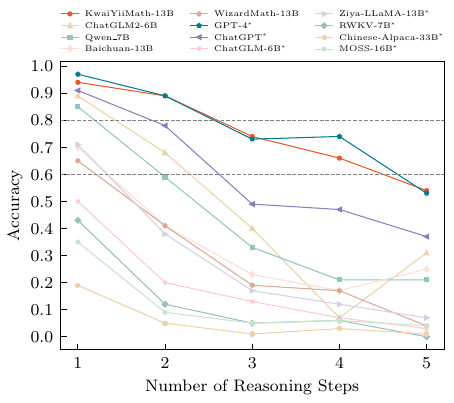}
     \caption{}
     \label{subgraph:b}
    \end{subfigure}

    \begin{subfigure}[b]{0.45\textwidth}
     \includegraphics[width=\textwidth]{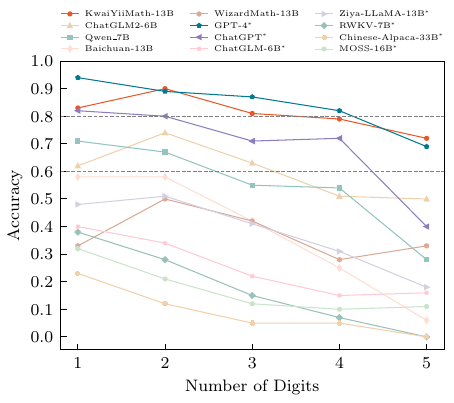}
     \caption{}
     \label{subgraph:c}
    \end{subfigure}
\hspace{-2mm}
    \begin{subfigure}[b]{0.45\textwidth}
     \includegraphics[width=\textwidth]{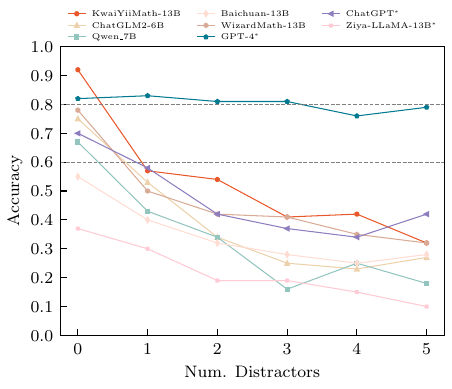}
     \caption{}
     \label{subgraph:d}
    \end{subfigure}

    \caption{(a) (b) (c): Average test accuracy against one of the problem complexity measures, including 
    grade, number of reasoning steps, and number of digits for each LLM. (d): Average test accuracy against the number of distractors on the distractor dataset. The character $*$ denotes that test accuracy is attained from paper~\cite{wei2023cmath} }
    \label{fig:figperformance}
\end{figure}
\subsection{Training and Evaluation Settings}
\paragraph{Training}
The meta prompt used in training of KwaiYiiMath is the version from Vicuna~\cite{vicuna2023}:
\textit{A chat between a curious user and an artificial intelligence assistant. The assistant gives helpful, detailed, and polite answers to the user's questions. USER: \{instruction\}. ASSISTANT: \{response\}}.

We follow standard fine-tuning hyperparameters: 3 epochs using AdamW~\cite{loshchilov2017decoupled} with $\beta_{1} = 0.9$, $\beta_{2} = 0.95$.
Without warmup steps, we set the initial learning rate to $4e-5$ and use the cosine learning rate decay strategy. 
The global batch size is set to $1024$ examples and texts longer than $2048$ tokens are trimmed.

\paragraph{Evaluation}
For evaluation on GSM8k~\cite{cobbe2021training}, we generate responses using the greedy decoding strategy.
For each sample in CMath~\cite{wei2023cmath} and KMath, we generate a single response from each baseline model using nucleus sampling~\cite{holtzman2019curious} with $top\_p = 0.9$ and a temperature of $\tau = 0.7$. 
We apply a repetition penalty of previously generated tokens with a hyperparameter of $1.01$~\cite{keskar2019ctrl}. 
We limit the maximum token length of output to 2048.

% Almost all the compared models generate two parts when answering a question: the chain-of-thought problem-solving process and the final answer.
Although all the compared models can generate the intermediate CoT process and final answer, we evaluate all LLMs on GSM8k~\cite{cobbe2021training} using few-shot CoT from~\cite {wei2022chain} for a fair comparison.
% For GSM8k, to facilitate comparison, models are evaluated using the few-shots chain-of-thought prompt from \cite{wei2022chain}.
We evaluate a solution as correct if the final answer matches the ground truth solution, independent of the quality of the CoT preceding it. 
To evaluate correctness, we parse the final answers and compare them using the SymPy library~\cite{meurer2017sympy}.
For CMath~\cite{wei2023cmath}, we use the code that is released with the data to evaluate the accuracy which only considers the correctness of the final answer.
For KMath, in order to evaluate the model results more comprehensively, we not only evaluate the correctness of the answers but also the correctness of the problem-solving process.
Specifically, we evaluate a solution as correct through human annotation if the answer is correct and the CoT problem-solving process is basically correct.
In order to eliminate the bias of human annotation, the correctness of each sample is first labeled by three different annotators, and then another quality assessment expert checks the labels.
The baseline models are evaluated in a zero-shot way since they have been aligned through SFT both for the CMath~\cite{wei2023cmath} and KMath.

\begin{wrapfigure}{R}{0.5\linewidth}
    % \centering
    % \vspace{-2mm}
    \begin{center}
    \includegraphics[width=0.8\linewidth]{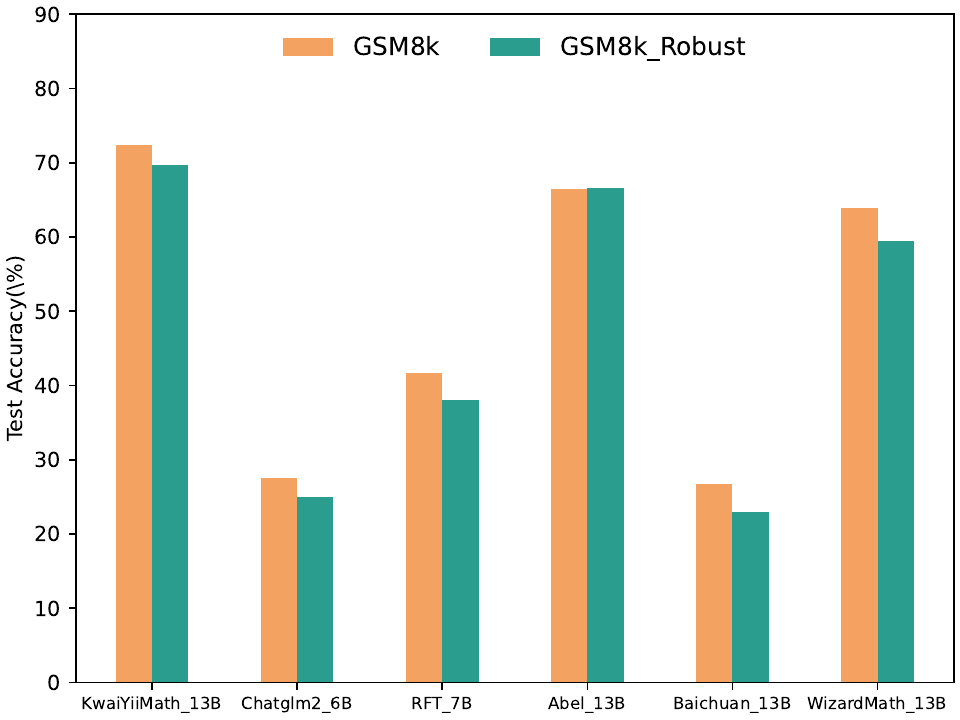}
    \end{center}
    \caption{Comparing the accuracy of the origin GSM8k and GSM8k\_Robust.}
    \label{fig:gsm8krobust}
    % \vspace{-10mm}
\end{wrapfigure}
\subsection{Main Results}
% In Table\ref{table:performance}, we report the main experimental results on three datasets.
% 在三个数据集上均超过同尺寸模型；在KMath上超过开源和闭源模型
Results on three datasets are shown in Table~\ref{table:performance}.
We observe that KwaiYiiMath outperforms the same size baseline LLMs on all benchmarks and also surpasses closed-source LLMs including ChatGPT and GPT4 on the KMath dataset, showing that finetuning on diverse and high-quality data is effective.
% 在CMath上
On the CMath dataset, KwaiYiiMath is close to GPT4 and also achieves a large improvement over other baseline models.
It shows that the KwaiYiiMath is not only effective on English mathematics problems but also on Chinese mathematics problems.
% 强化后进一步提升
% It can also be found that 
Meanwhile, KwaiYiiMath-HPA achieves improvement over KwaiYiiMath on three benchmarks, showing the effectiveness of the human preference alignment process.

% \footnotetext{We estimated the corresponding values from the figures of ~\cite{wei2023cmath}, so slight gaps may exist.}

\subsection{Fine-grained and Robustness Results}
In this subsection, we investigate the performance of the LLMs on mathematical problems of varying complexity.
% The CMath~\cite{wei2023cmath} data set provides three dimensions of math question complexity differentiation, namely primary school grade of question, the number of digits that an LLM needs to manipulate, and the number of reasoning steps that an LLM needs to carry out in order to solve a problem.

\paragraph{Fine-grained Results}
The CMath~\cite{wei2023cmath} dataset provides the primary school grade corresponding to the question, which can indicate the comprehensive complexity of the question. 
In addition, two dimensions are provided to more intuitively represent the complexity of the question, namely the number of digits that an LLM needs to manipulate, and the number of reasoning steps that an LLM needs to carry out in order to solve a problem.
% 一个维度表示综合难度, 两个维度标注计算和推理复杂度
% 不同的维度的含义, 以及他们可能对于效果的影响.
Intuitively, problems with higher arithmetic complexity or reasoning complexity should be harder to solve, resulting in lower accuracy.

In Figure \ref{subgraph:a}, a distinct downward trend in accuracy is evident, signifying that the performance of all models declines as the complexity increases.
GPT-4 and KwaiYiiMath are the only two models that achieve success (accuracy exceeding 60\%) in math tests across all six elementary school grades and achieve high accuracy (exceeding 80\%) in tests for grades 1 to 4.
The performance of KwaiYiiMath is very close to GPT-4, and even slightly outperforms GPT-4 in some grades.
Following GPT-4 and KwaiYiiMath, ChatGPT, ChatGLM2-6b, and Qwen-7B demonstrate success in tests for grades 1 to 4, but encounter difficulties in grades 5 and 6 (accuracy under 60\%). 

From Figure~\ref{subgraph:b} and~\ref{subgraph:c}, it can be found that all models’ performance declines as either of the problem complexity measures augments.
Judged from the downward slopes of the plots, it is pertinent to say that the reasoning complexity of the problem has generally a larger impact than the arithmetic complexity.

\paragraph{Robustness Results}
% 先写鲁棒性较好的部分, 再写不足的部分，以及对应的分析
The robustness experiment consists of two parts: adversarial evaluation on the GSM8k robust dataset and the CMath distractor dataset. 
The GSM8k robust dataset is a dataset released by~\cite{abel} that was established based on the GSM8k dataset. 
~\cite{abel} randomly modified the numbers within the questions of the GSM8k test set, without altering any other information in the questions, using GPT-4. 
The GSM8k robust dataset can be used to evaluate whether the models overfit the training data, making the models susceptible to out-of-distribution testing samples.
Fiture~\ref{fig:gsm8krobust} shows the performance of KwaiYiiMath on GSM8k robust dataset\footnote{The results of RFT, Abel, and WizardMath are attained from ~\cite{abel}}.
It can be found that there is a slight decrease in the performance of KwaiYiiMath, which demonstrates that the KwaiYiiMath also has strong robustness out-of-distribution testing samples.

% 鲁棒性的维度
The  CMath distractor dataset is a small set released with CMath~\cite{wei2023cmath} for testing the robustness of the model to irrelevant information in the question.
The CMath distractor dataset contains 360 questions: 60 seed questions and 5 noisy versions of each seed question with varying numbers of distractors, from 1 to 5.
The performance of models on the Cmath distractor dataset is plotted in Figure~\ref{subgraph:d}.
From the figure, it can be observed that we observe that the performance of all LLMs, with the exception of GPT-4, drops drastically as the number of distractors increases.
% 展示了一定的鲁棒性，在所有模型中位列第二，且全部大于等于80%.
KwaiYiiMath shows a slight ability of robustness, with the overall effect ranking second among all models, and the accuracy on the 5 distractors subset is still greater than 40\%.
However, KwaiYiiMath also suffers an accuracy drop of nearly 50\% for problems augmented with merely three distractors.
We guess the performance decrease is because the training data used by the KwaiYiiMath are relatively clean questions, that is, there is no distractor information in the questions.
Therefore, when distractor information appears in the question, especially information that is very similar to the origin question to be solved, the model will generate many useless steps to solve the problem, resulting in the final answer being wrong.
This is also an important ability that needs to be improved in the future.
% We did a preliminary analysis to understand why KwaiYiiMath does not perform well with distractor information in the question.

% 这里放GSM8k鲁棒性测试集的表格或柱状图，如果可以的话
% \begin{wrapfigure}{r}
%     \begin{center}
%         \includegraphics{figs/llm_robust.pdf}
%         \caption{Comparing the accuracy of the origin GSM8k and GSM8k_Robust.}
%         \label{fig:gsm8krobust}
%     \end{center}
% \end{wrapfigure}

\section{Conclusions}
In this report, we introduce the KwaiYiiMath which is fine-tuned from KwaiYiiBase to tackle mathematical problems.
By utilizing a large amount of high-quality mathematical data to perform the human alignment process, including SFT and RLHF, we have greatly enhanced the mathematical reasoning capabilities of KwaiYiiMath.
Experimental results also show that KwaiYiiMath outperforms many open-source models in similar sizes by a large margin and is approaching GPT-4 on three mathematical benchmarks including English and Chinese.
Experimental results on relevant robust data sets also show that the KwaiYiiMath is robust to the disturbed information in question.

Although our work has achieved results close to GPT-4 on relevant benchmarks, in fact, there is still a big gap in a wider range of tasks.
In the future, we hope to further explore the methods to improve the mathematical reasoning capabilities of LLMs and the intrinsic mechanism behind data augmentation for LLMs.

\bibliography{custom}
\bibliographystyle{iclr2024_conference}

\appendix
\section{Appendix}
\subsection{Case Study}

% \begin{tabular}{ l|c } 
%  \hline
% \multicolumn{2}{l}{\bf{Instruction:} \begin{CJK*}{UTF8}{gbsn} 给直径为0.75米的水缸做一个木盖，木盖的直径比缸口直径大5厘米，这个木盖的面积\end{CJK*} } \\
% &\begin{CJK*}{UTF8}{gbsn}是多少平方米？ \end{CJK*}  \\

%  \hline
% \end{tabular}

% GSM8k的两个case
% case1

\begin{table}[h!]
\caption{A comparison case on KwaiYiiMath, ChatGPT, and GPT4 of the GSM8k dataset. All three models answer correctly, with GPT4 giving the most concise answer that only included the core calculation steps.}
\begin{tabular}{ c |l } 
 \hline
 \multicolumn{2}{l}{\textbf{Instruction:} While on vacation in Bali, Thea bought a hat from a craftsman worth \$70. } \\
 \multicolumn{2}{l}{If she gave the craftsman four \$20 bills, how much change did she get? }\\

 \hline
 \multicolumn{2}{l}{\textbf{Ground Truth Answer:} 10} \\
 \hline
 KwaiYiiMath & First, we need to find out how much money Thea gave to the craftsman in total.\\
             & Since she gave him four \$20 bills, that means she gave him 4 * \$20 = \$80. \\
             & Now, we know that the hat cost \$70, so to find out how much change she got, \\
             & we simply subtract the cost of the hat from the total amount of money she gave: \\ 
             & \$80 - \$70 = \$10. So, Thea got \$10 in change. The answer is 10.\\ 
  \hline
  ChatGPT & The total amount Thea paid the craftsman with four \$20 bills is: 4 x \$20 = \$80 \\
             & Since the hat only costs \$70, Thea got a change of: \$80 - \$70 = \$10 \\
             & Therefore, Thea got a change of \$10. \\
 \hline
 GPT4 & Thea gave the craftsman \$20 x 4 = \$$<<20*4=80>>$80. \\
      & So, she got \$80 - \$70 = \$$<<80-70=10>>$10 change.\\
      & Answer: \boxed{10}.\\
 \hline
\end{tabular}
\end{table}

% case2
\begin{table}[h!]
\caption{A comparison case on KwaiYiiMath, ChatGPT, and GPT4 of the GSM8k dataset. KwaiYiiMath and GPT4 answer correctly, while ChatGPT answers incorrectly.}
\begin{tabular}{ c |l } 
 \hline
 \multicolumn{2}{l}{\textbf{Instruction:} Robert wants to buy a bookshelf and a sofa and has received two offers.} \\
 \multicolumn{2}{l}{Furniture United's offer includes a \$1,350 advance payment and 6 monthly installments of \$350 each.  }\\
 \multicolumn{2}{l}{Cozy Homes' offer includes a \$1,100 advance payment and 9 monthly installments of \$250 each.  }\\
 \multicolumn{2}{l}{What is the cost difference between the two offers? }\\
 \hline
 \multicolumn{2}{l}{\textbf{Ground Truth Answer:} 100} \\
 \hline
 KwaiYiiMath & First, we need to calculate the total cost of each offer.\\
             & For Furniture United, the total cost is the sum of the advance payment and the \\ 
             & total amount paid in monthly installments. For Cozy Homes, the total cost is \\
             & the sum of the advance payment and the total amount paid in monthly installments.\\
             & Then, we subtract the lower total cost from the higher total cost to find the cost difference.\\ 
             & Furniture United: \$1,350 + (6 * \$350) = \$1,350 + \$2,100 = \$3,450\\ 
             & Cozy Homes: \$1,100 + (9 * \$250) = \$1,100 + \$2,250 = \$3,350\\ 
             & Cost difference: \$3,450 - \$3,350 = \$100.  The answer is 100.\\
             
  \hline
  ChatGPT & The total cost of Furniture United's offer can be calculated as: \\
             & \$1,350 advance payment + 6 x \$350 monthly installments = \$3,150 \\
             & The total cost of Cozy Homes' offer can be calculated as:\\
             & \$1,100 advance payment + 9 x \$250 monthly installments = \$3,350\\
             & To find the cost difference, we subtract the cost of Furniture United's offer from the \\
             & cost of Cozy Homes' offer: \\
             & \$3,350 (Cozy Homes' offer) - \$3,150 (Furniture United's offer) = \$200 \\
             & Therefore, the cost difference between the two offers is \$200. \\
             & Furniture United's offer is \$200 cheaper than Cozy Homes' offer. \\
   
 \hline
 GPT4 & For Furniture United, the total cost of the monthly installments is \\
      & \$350 * 6 = \$$<<350*6=2100>>$2,100. So, their total offer's cost is \\
      & \$1,350 (advance payment) + \$2,100 (installments) = \$$<<1350+2100=3450>>$3,450. \\
      & For Cozy Homes, the total cost of the monthly installments \\
      & is \$250 * 9 = \$$<<250*9=2250>>$2,250. So, their total offer's cost is \\
      & \$1,100 (advance payment) + \$2,250 (installments) = \$$<<1100+2250=3350>>$3,350.  \\
      & The cost difference between the two offers is \$3,450 (Furniture United) \\
      & - \$3,350 (Cozy Homes) = \$$<<3450-3350=100>>$100. Answer: \boxed{100}. \\
 \hline
\end{tabular}
\end{table}

\begin{table}[h!]
\caption{A comparison case on KwaiYiiMath, ChatGPT, and GPT4 of CMath dataset. KwaiYiiMath and GPT4 answer correctly, while ChatGPT answers incorrectly.}
\begin{tabular}{ c|l } 
 \hline
 \multicolumn{2}{l}{\textbf{Instruction:} \begin{CJK*}{UTF8}{gbsn} 给直径为0.75米的水缸做一个木盖，木盖的直径比缸口直径大5厘米，\end{CJK*}} \\
 \multicolumn{2}{l}{\begin{CJK*}{UTF8}{gbsn}这个木盖的面积是多少平方米？ \end{CJK*}}\\
 \hline
 \multicolumn{2}{l}{\textbf{Ground Truth Answer:} 0.5024} \\
 \hline
 KwaiYiiMath & \begin{CJK*}{UTF8}{gbsn} 首先，将木盖直径与水缸直径相差的5厘米转换为米，即0.05米。\end{CJK*}\\
             & \begin{CJK*}{UTF8}{gbsn} 那么，木盖的直径为0.75米+0.05米=0.8米。\end{CJK*}\\
             & \begin{CJK*}{UTF8}{gbsn} 接下来，计算木盖的半径，即0.8米÷2=0.4米。\end{CJK*}\\
             & \begin{CJK*}{UTF8}{gbsn} 最后，利用圆的面积公式计算木盖的面积：\end{CJK*}\\ 
             & \begin{CJK*}{UTF8}{gbsn} 3.14×(0.4米)²=3.14×0.16=0.5024平方米。\end{CJK*}\\ 
             & \begin{CJK*}{UTF8}{gbsn} 因此，这个木盖的面积是0.5024平方米。\end{CJK*} \\
  \hline
  ChatGPT & \begin{CJK*}{UTF8}{gbsn} 水缸的直径为0.75米，则半径为 0.75m / 2 = 0.375米。\end{CJK*}\\
             & \begin{CJK*}{UTF8}{gbsn} 木盖的直径比缸口直径大5厘米，即比缸口半径大0.05米，因此木盖的半径为：\end{CJK*}\\
             & \begin{CJK*}{UTF8}{gbsn} 0.375米 + 0.05米 = 0.425米。\end{CJK*}\\
             & \begin{CJK*}{UTF8}{gbsn} 那么，木盖的面积为：π × (0.425米)² - π × (0.375米)² ≈ 0.155平方米\end{CJK*}\\ 
             & \begin{CJK*}{UTF8}{gbsn} 因此，这个木盖的面积约为0.155平方米。\end{CJK*}\\ 
 \hline
 GPT4 & \begin{CJK*}{UTF8}{gbsn} 木盖的直径比缸口直径大5厘米，所以木盖的直径是：0.75米 + 0.05米 = 0.8米。\end{CJK*}\\
             & \begin{CJK*}{UTF8}{gbsn} 木盖的半径为：0.8米 ÷ 2 = 0.4米。\end{CJK*}\\
             & \begin{CJK*}{UTF8}{gbsn} 木盖的面积为圆的面积公式：π × r²，其中r为半径，π ≈ 3.14。\end{CJK*}\\
             & \begin{CJK*}{UTF8}{gbsn} 所以木盖的面积为：3.14 × (0.4米)² = 3.14 × 0.16平方米 ≈ 0.5024平方米。 \end{CJK*}\\ 
 \hline
\end{tabular}
\end{table}

\begin{table}[h!]
\caption{A comparison case on KwaiYiiMath, ChatGPT, and GPT4 of CMath dataset. All three models use a series of equations to solve this question. KwaiYiiMath and GPT4 answer correctly, while ChatGPT answers incorrectly.}
\begin{tabular}{ c|l } 
 \hline
 \multicolumn{2}{l}{\textbf{Instruction:} \begin{CJK*}{UTF8}{gbsn} 妈妈和李阿姨一起到超市购物，妈妈买了5千克大米和4千克面粉共付29.2元，\end{CJK*}} \\
 \multicolumn{2}{l}{\begin{CJK*}{UTF8}{gbsn}李阿姨也买了同样的4千克大米和5千克面粉则共付28.4元，每千克大米多少钱？ \end{CJK*}}\\
 \hline
 \multicolumn{2}{l}{\textbf{Ground Truth Answer:} 3.6} \\
 \hline
 KwaiYiiMath & \begin{CJK*}{UTF8}{gbsn} 设每千克大米的价格为x元，每千克面粉的价格为y元。\end{CJK*}\\
             & \begin{CJK*}{UTF8}{gbsn} 根据题目信息，我们可以得到以下两个方程：\end{CJK*}\\
             & \begin{CJK*}{UTF8}{gbsn} 5x + 4y = 29.2\end{CJK*}\\
             & \begin{CJK*}{UTF8}{gbsn} 4x + 5y = 28.4\end{CJK*}\\ 
             & \begin{CJK*}{UTF8}{gbsn} 我们可以用消元法来解这个方程组。\end{CJK*}\\ 
             & \begin{CJK*}{UTF8}{gbsn} 首先，我们将第一个方程乘以4，第二个方程乘以5，得到：\end{CJK*} \\
             & \begin{CJK*}{UTF8}{gbsn} 20x + 16y = 116.8\end{CJK*} \\
             & \begin{CJK*}{UTF8}{gbsn} 20x + 25y = 142\end{CJK*} \\
             & \begin{CJK*}{UTF8}{gbsn} 接下来，我们将第二个方程从第一个方程中减去，得到：\end{CJK*} \\
             & \begin{CJK*}{UTF8}{gbsn} 9y = 25.2\end{CJK*} \\
             & \begin{CJK*}{UTF8}{gbsn} 然后，我们求解y:\end{CJK*} \\
             & \begin{CJK*}{UTF8}{gbsn} y = 25.2 / 9\end{CJK*} \\
             & \begin{CJK*}{UTF8}{gbsn} y = 2.8\end{CJK*} \\
             & \begin{CJK*}{UTF8}{gbsn} 现在我们知道每千克面粉的价格是2.8元，我们可以将y的值代入第一个方程，求解x \end{CJK*} \\
             & \begin{CJK*}{UTF8}{gbsn} 5x + 4(2.8) = 29.2\end{CJK*} \\
             & \begin{CJK*}{UTF8}{gbsn} 5x + 11.2 = 29.2\end{CJK*} \\
             & \begin{CJK*}{UTF8}{gbsn} 5x = 18\end{CJK*} \\
             & \begin{CJK*}{UTF8}{gbsn} x = 18 / 5\end{CJK*} \\
             & \begin{CJK*}{UTF8}{gbsn} x = 3.6\end{CJK*} \\
             & \begin{CJK*}{UTF8}{gbsn} 所以每千克大米的价格是3.6元。\end{CJK*} \\
             
  \hline
  ChatGPT & \begin{CJK*}{UTF8}{gbsn} 设每千克大米的价格为 x 元，则每千克面粉的价格为：\end{CJK*}\\
             & \begin{CJK*}{UTF8}{gbsn} 29.2元 - 5千克 × x元/千克 ÷ 4千克 = 28.4元 - 4千克 × x元/千克 ÷ 5千克 \end{CJK*}\\
             & \begin{CJK*}{UTF8}{gbsn} 简化可得：7.3元/千克 - 5/4 x元/千克 = 5.68元/千克 - x元/千克 \end{CJK*}\\
             & \begin{CJK*}{UTF8}{gbsn} 移项合并同类项可得：2.62元/千克 = 1/4 x元/千克 \end{CJK*}\\ 
             & \begin{CJK*}{UTF8}{gbsn} 即：x元/千克 = 2.62 × 4 = 10.48元/千克 \end{CJK*}\\ 
             & \begin{CJK*}{UTF8}{gbsn} 因此，每千克大米的价格为10.48元。 \end{CJK*}\\
 \hline
 GPT4 & \begin{CJK*}{UTF8}{gbsn} 设每千克大米价格为x元，每千克面粉价格为y元，可以得到以下方程组：\end{CJK*}\\
             & \begin{CJK*}{UTF8}{gbsn} 5x + 4y = 29.2（1）\end{CJK*}\\
             & \begin{CJK*}{UTF8}{gbsn} 4x + 5y = 28.4（2）\end{CJK*}\\
             & \begin{CJK*}{UTF8}{gbsn} 为了求解这个方程组，我们可以将两个方程相加：\end{CJK*}\\ 
             & \begin{CJK*}{UTF8}{gbsn} (5x + 4y) + (4x + 5y) = 29.2 + 28.4 \end{CJK*}\\
             & \begin{CJK*}{UTF8}{gbsn} 9x + 9y = 57.6 \end{CJK*}\\
             & \begin{CJK*}{UTF8}{gbsn}  x + y = 6.4 （3） \end{CJK*}\\
             & \begin{CJK*}{UTF8}{gbsn} 然后将方程（1）减去方程（2）： (5x + 4y) - (4x + 5y) = 29.2 - 28.4 x - y = 0.8 （4）\end{CJK*}\\
             & \begin{CJK*}{UTF8}{gbsn} 现在我们可以解这个方程组来找到每千克大米的价格： \end{CJK*}\\
             & \begin{CJK*}{UTF8}{gbsn} 将方程（3）和（4）相加得到： (x + y) + (x - y) = 6.4 + 0.8 2x = 7.2 \end{CJK*}\\
             & \begin{CJK*}{UTF8}{gbsn} 所以每千克大米价格为: x = 7.2 ÷ 2 = 3.6元 \end{CJK*}\\
\hline
\end{tabular}
\end{table}

% KMath中的一个case
% \begin{table}[h!]
% \caption{A comparison case on KwaiYiiMath, ChatGPT, and GPT4 of KMath dataset. ChatGPT and GPT4 answer incorrectly, while KwaiYiiMath answers correctly.}
% \begin{tabular}{ l|l } 
%  \hline
%  \multicolumn{2}{l}{\textbf{Instruction:} \begin{CJK*}{UTF8}{gbsn} 一百一百地数，800后面第二个数是（ ）。A:600; B:700; C:900; D:1000\end{CJK*}} \\
%  \hline
%  \multicolumn{2}{l}{\textbf{Ground Truth Answer:} D} \\
%  \hline
%  KwaiYiiMath & \begin{CJK*}{UTF8}{gbsn} 首先，我们需要找到在数到800之后，下一个一百的数是900。\end{CJK*}\\
%              & \begin{CJK*}{UTF8}{gbsn} 然后，在数到900之后，下一个一百数的数是1000。\end{CJK*}\\
%              & \begin{CJK*}{UTF8}{gbsn} 因此，800后面第二个数是1000。所以答案是D\end{CJK*}\\
%   \hline
%   ChatGPT & \begin{CJK*}{UTF8}{gbsn} 答案是B:700。\end{CJK*}\\
%  \hline
%  GPT4 & \begin{CJK*}{UTF8}{gbsn} 一百一百地数，我们可以得到这样一个数列：\end{CJK*}\\
%              & \begin{CJK*}{UTF8}{gbsn} 100，200，300，400，500，600，700，800，900，1000...。 \end{CJK*}\\
%              & \begin{CJK*}{UTF8}{gbsn} 从800开始，向后数第二个数是900，所以答案是C。\end{CJK*}\\
%  \hline
% \end{tabular}
% \end{table}

\begin{table}[h!]
\caption{A comparison case on KwaiYiiMath, ChatGPT, and GPT4 of KMath dataset. ChatGPT and GPT4 answer incorrectly, while KwaiYiiMath answers correctly.}
\begin{tabular}{ c|l } 
 \hline
 \multicolumn{2}{l}{\textbf{Instruction:} \begin{CJK*}{UTF8}{gbsn} 一根电线，第一次用去480米，比第二次用去的1.5倍少60米，第二次用去电线多少米？\end{CJK*}} \\
 \hline
 \multicolumn{2}{l}{\textbf{Ground Truth Answer:} 360} \\
 \hline
 KwaiYiiMath & \begin{CJK*}{UTF8}{gbsn} 设第二次用去的电线长度为x米，根据题意可得方程: \end{CJK*}\\
             & \begin{CJK*}{UTF8}{gbsn} 480 = 1.5x - 60。解这个方程，我们可以得到：\end{CJK*}\\
             & \begin{CJK*}{UTF8}{gbsn} 1.5x - 60 = 480 \end{CJK*}\\
             & \begin{CJK*}{UTF8}{gbsn} 1.5x = 540 \end{CJK*}\\
             & \begin{CJK*}{UTF8}{gbsn} x = 360 \end{CJK*}\\
             & \begin{CJK*}{UTF8}{gbsn} 所以，第二次用去的电线长度为360米。 \end{CJK*}\\
  \hline
  ChatGPT & \begin{CJK*}{UTF8}{gbsn} 设第二次用去的电线长度为x，则第一次用去的电线长度为2x-60 \end{CJK*}\\
             & \begin{CJK*}{UTF8}{gbsn} （因为比第二次用去的1.5倍少60米） \end{CJK*}\\
             & \begin{CJK*}{UTF8}{gbsn} 根据题意得：第一次用去的电线长度 + 第二次用去的电线长度 = 总电线长度 \end{CJK*}\\
             & \begin{CJK*}{UTF8}{gbsn} 即：2x-60 + x = 480 \end{CJK*}\\
             & \begin{CJK*}{UTF8}{gbsn} 化简得：3x = 540 \end{CJK*}\\
             & \begin{CJK*}{UTF8}{gbsn} 解方程得：x = 180 \end{CJK*}\\
             & \begin{CJK*}{UTF8}{gbsn} 所以第二次用去的电线长度为180米。 \end{CJK*}\\
 \hline
 GPT4 & \begin{CJK*}{UTF8}{gbsn} 设第二次用去的电线长度为x米，\end{CJK*}\\
             & \begin{CJK*}{UTF8}{gbsn} 则第一次用去的电线长度为1.5x-60米。 \end{CJK*}\\
             & \begin{CJK*}{UTF8}{gbsn} 根据题意，两次用去的电线总长度为480米，可以得到方程： \end{CJK*} \\
             & \begin{CJK*}{UTF8}{gbsn} 1.5x - 60 + x = 480 \end{CJK*} \\
             & \begin{CJK*}{UTF8}{gbsn} 解这个方程，我们可以得到：\end{CJK*} \\
             & \begin{CJK*}{UTF8}{gbsn} 2.5x = 540 \end{CJK*} \\
             & \begin{CJK*}{UTF8}{gbsn} x = 216 \end{CJK*} \\
             & \begin{CJK*}{UTF8}{gbsn} 所以，第二次用去的电线长度为216米。 \end{CJK*} \\
 \hline
\end{tabular}
\end{table}

\begin{table}[h!]
\caption{A additional comparison case on KwaiYiiMath, ChatGPT, and GPT4 of calculus field.}
\begin{tabular}{ c|l } 
 \hline
 \multicolumn{2}{l}{\textbf{Instruction:} \begin{CJK*}{UTF8}{gbsn} 求g(x)=exp(x\^{}2+1)对x的微分 \end{CJK*}} \\
 \hline
 KwaiYiiMath & \begin{CJK*}{UTF8}{gbsn} 要计算g(x) = exp(x\^{}2+1)关于x的微分，我们需要使用链式法则。\end{CJK*}\\
             & \begin{CJK*}{UTF8}{gbsn} 首先，我们求导exp(x\^{}2+1)关于x的导数。\end{CJK*}\\
             & \begin{CJK*}{UTF8}{gbsn} exp(x\^{}2+1)关于x的导数可以表示为： \end{CJK*}\\
             & \begin{CJK*}{UTF8}{gbsn} d/dx(exp(x\^{}2+1)) = d/dx(exp(x\^{}2+1)) * d/dx(x\^{}2+1) \end{CJK*}\\
             & \begin{CJK*}{UTF8}{gbsn} 这里，d/dx(x\^{}2+1) = 2x，因为x\^{}2的导数是2x。 \end{CJK*}\\
             & \begin{CJK*}{UTF8}{gbsn} 所以，d/dx(exp(x\^{}2+1)) = exp(x\^{}2+1) * 2x \end{CJK*}\\
             & \begin{CJK*}{UTF8}{gbsn} 现在我们求导g(x) = exp(x\^{}2+1)关于x的导数， \end{CJK*}\\
             & \begin{CJK*}{UTF8}{gbsn} 就是求导d/dx(exp(x\^{}2+1)) * d/dx(x\^{}2+1)： \end{CJK*}\\
             & \begin{CJK*}{UTF8}{gbsn} dg(x)/gx = exp(x\^{}2+1) * 2x \end{CJK*}\\
              & \begin{CJK*}{UTF8}{gbsn} 这就是g(x)关于x的微分。 \end{CJK*}\\
             
  \hline
  ChatGPT & \begin{CJK*}{UTF8}{gbsn} g'(x) = d/dx(exp(x\^{}2+1)) \end{CJK*}\\
             & \begin{CJK*}{UTF8}{gbsn} 使用链式法则，将指数函数部分的导数分离出来并乘以内部函数的导数： \end{CJK*}\\
             & \begin{CJK*}{UTF8}{gbsn} g'(x) = exp(x\^{}2+1) * d/dx(x\^{}2+1) \end{CJK*}\\
             & \begin{CJK*}{UTF8}{gbsn} 对于 x\^{}2+1 部分，也使用链式法则： \end{CJK*}\\
             & \begin{CJK*}{UTF8}{gbsn} d/dx(x\^{}2+1) = 2x * 1 \end{CJK*}\\
             & \begin{CJK*}{UTF8}{gbsn} 将其代入原式得：g'(x) = exp(x\^{}2+1) * 2x \end{CJK*}\\
             & \begin{CJK*}{UTF8}{gbsn} 因此，g(x) 对 x 的微分为：g'(x) = 2x * exp(x\^{}2+1) \end{CJK*}\\
 \hline
 GPT4 & \begin{CJK*}{UTF8}{gbsn} 为了求g(x)=exp(x\^{}2+1)关于x的微分，我们采用链式法则求导。\end{CJK*}\\
             & \begin{CJK*}{UTF8}{gbsn} 链式法则表示为：假设y=f(u)和u=g(x)是可微函数，则复合函数y=f(g(x))的导数为： \end{CJK*}\\
             & \begin{CJK*}{UTF8}{gbsn} (dy/dx) = (dy/du) * (du/dx) \end{CJK*} \\
             & \begin{CJK*}{UTF8}{gbsn} 这里，令u = x\^{}2 + 1，那么g(x) = exp(u)。 \end{CJK*} \\
             & \begin{CJK*}{UTF8}{gbsn} 首先计算du/dx： du/dx = d(x\^{}2 + 1)/dx = 2x \end{CJK*} \\
             & \begin{CJK*}{UTF8}{gbsn} 然后计算dy/du： dy/du = d(exp(u))/du = exp(u) \end{CJK*} \\
             & \begin{CJK*}{UTF8}{gbsn} 现在我们可以计算dy/dx： dy/dx = (dy/du) * (du/dx) = exp(u) * 2x = exp(x\^{}2 + 1) * 2x \end{CJK*} \\
             & \begin{CJK*}{UTF8}{gbsn} 所以，g(x)=exp(x\^2+1)关于x的微分为：2x * exp(x\^{}2 + 1)。 \end{CJK*} \\
 \hline
\end{tabular}
\end{table}

\end{document}